%% file: vmt.tex
\documentclass{article}
\usepackage[T1]{fontenc} % add special characters (e.g., umlaute)
\usepackage[utf8]{inputenc} % set utf-8 as default input encoding
\usepackage{amssymb}
\usepackage{ismir,amsmath,cite,url}
\usepackage{graphicx}
\usepackage{color}

\usepackage[bookmarks=false,hidelinks]{hyperref}
\usepackage{amssymb}
\usepackage{subfig}
\usepackage{booktabs}
\usepackage{verbatim}
\makeatletter
\def\verbatim@font{\normalfont \ttfamily \bfseries \hyphenchar \font \m@ne \@noligs}
\makeatother

\usepackage{xspace}

\renewcommand{\paragraph}[1]{\textbf{#1}}

\DeclareMathOperator{\Attention}{Attention}
\DeclareMathOperator{\AvgPool}{AvgPool}
\DeclareMathOperator{\FFN}{FFN}
\DeclareMathOperator{\LayerNorm}{LayerNorm}
\DeclareMathOperator{\LeakyReLu}{LeakyReLu}
\DeclareMathOperator{\SoftMax}{SoftMax}

\newcommand{\SeqSeq}{\textsc{Seq2Seq}\xspace}

\title{InverseMV: Composing Piano Scores with a Convolutional Video-Music Transformer}

\twoauthors
 {Chin-Tung Lin} {Institute of Computer Science \\ National Chengchi University \\ {\tt lctung5@gmail.com}}
 {Mu Yang} {Institute of Information Science \\ Acdemia Sinica \\ {\tt emfomy@gmail.com}}

\begin{document}

\maketitle
%
\input{content/0-abstract}
\input{content/1-introduction}
\input{content/2-data}
\input{content/3-model}
\input{content/4-experiment}
\input{content/5-related}
\input{content/6-conclusion}

% For bibtex users:
\bibliography{vmt}

\end{document}

%% file: content/0-abstract.tex
%!TEX root = ../vmt.tex

\begin{abstract}
Many social media users prefer consuming content in the form of videos rather than text.
However, in order for content creators to produce videos with a high click-through rate, much editing is needed to match the footage to the music.
This posts additional challenges for more amateur video makers.
Therefore, we propose a novel attention-based model \textbf{VMT} (Video-Music Transformer) that automatically generates piano scores from video frames.
Using music generated from models also prevent potential copyright infringements that often come with using existing music.
To the best of our knowledge, there is no work besides the proposed VMT that aims to compose music for video. Additionally, there lacks a dataset with aligned video and symbolic music.
We release a new dataset composed of over 7 hours of piano scores with fine alignment between pop music videos and MIDI files. 
We conduct experiments with human evaluation on VMT, \SeqSeq model (our baseline), and the original piano version soundtrack. VMT achieves consistent improvements over the baseline on music smoothness and video relevance.
In particular, with the relevance scores and our case study, our model has shown the capability of multimodality on frame-level actors' movement for music generation. Our VMT model, along with the new dataset, presents a promising research direction toward composing the matching soundtrack for videos. We have released our code at \url{https://github.com/linchintung/VMT}.

\end{abstract}

%% file: content/1-introduction.tex
%!TEX root = ../vmt.tex

\section{Introduction} \label{sec:introduction}

With the blooming of mobile technologies, daily updates of personal news from celebrities, athletes and people of influence have become a culture phenomenon.
Most content posted on social media platforms such as Facebook and Twitter has gradually drifted from pure text to multi-media such as videos and pictures.
%% for good reason?
According to Twitter Marketing UK, tweets with videos, which are the most shared media type, are retweeted six times more than those with photos, and three times more than those with GIFs.

%%???
The most popular videos on social media, shared and liked by millions, are often accompanied with background music carefully edited to complement the video.
Although with the improved video capacities of modern mobile devices, filming high quality videos has become much easier, producing suitable background music to match the videos remains a challenging task.
Copyrighted music created by human experts might not synchronize with person moves in the video.
%% time consuming to train experts
%% vvv questionable stance...
Even if we want to use kinds of music created by experts, copyright is a problem that prevents us from doing so.

To tackle the above problems, an automatic music generation system is helpful.
In addition to preventing breaching copyright when using existing music, automatic music generation for video makes editing a multimedia post easier and more efficient.
To this end, we propose a novel model to generate symbolic piano music for video. To the best of our knowledge, there is no previous work dealt with generating music for video.
Considering that the goal of the task is to generate music from a given video, video music generation (VMG) is challenging in a few aspects.

Firstly, there is no existing dataset that we can train a suite of models on VMG task.
Previous work on background music recommendation~\cite{kuo2005emotion, han2010music, lin2015emv} is to extract emotion from facial, voice, physiological signals and text and
then affect them on music elements including melody, rhythm, tempo and text. Hence, their work focus on modeling overall features rather than local features (ex. people dancing frame in units of music notes). We expect our generated music formed up by piano notes which suits every video frame. Therefore, we release a new dataset, \emph{MIDI and Video Edited for Synchronous Piano Notes and Music Videos}, composed of over 7 hours of piano scores with fine alignment between pop music videos and MIDI tracks.

Secondly, there is no previous work dealing with the video-to-music generation (VMG) task. The task of VMG needs a mechanism to model multi-modal video and song. We are inspired by the success of video caption generation, but unlike text, music is an artistic creation that can't be interpreted consistently.
Moreover, music sequences have more diversity and long-term structure rather than caption sequences. Recent advances in attention-based neural networks with properties of handling long-term structures have made it possible to train the music generation model.
Thus, we propose Video Music Transformer (VMT), a novel attention-based multi-modal model to generate music for given videos. Experiment results show VMT substantially outperform Seq2Seq on both music smoothness and video relevance.

Our contributions can be summarized as follows:
\begin{itemize}
\item We propose VMT, a novel attention-based multi-modal model to generate music for a given video.
\item With the lack of training data of alignment between video and MIDI notes, we release a new dataset composed of over 7 hours of videos, including over 2500 videos with aligned MIDI files.
\end{itemize}

%% file: content/2-data.tex
%!TEX root = ../vmt.tex

\section{Dataset}

One crucial challenge of this work lies in the lack of training data.
At the time of writing, there is no existing dataset for video to MIDI music generation.
Music Transformer~\cite{huang2018music} released a rich classical piano MIDI dataset. 
However, since most classical music are simply accompanied by recordings of the live performance.
% Therefore, we decide to create a new dataset for this task.
In this paper, we created a new dataset for the task of video to music generation.
% A massive classical piano MIDI dataset from Music Transformer~\cite{huang2018music} allows us to pre-train a model of piano MIDI generation.
% With means to generate piano MIDI, we only need to generate a relatively small dataset in order to teach our model the relation between music and video.

%------------------------------------------------------------------------------%

\subsection{Data Collection}

We collected pop music as the training data for our (task) for the following reasons:
\begin{itemize}
    \item In many music video fragments, the singer's moves are synchronized with the groove of the music.
    \item Pop music videos are plentiful with camera angles, shots, and movements, which offer benefits to learning variety of videos and corresponding music.
\end{itemize}
Although there is an existing classing music MIDI dataset, it is hard to find corresponding music there are seldom music videos for classical music songs.
However, most pop music songs have official music videos, which are usually filmed by famous directors and edited by experts.
These videos usually fit music's emotion and representation with compare to other kinds of music.

We collected the most famous music videos from YouTube channels
Vevo, the world's largest all-premium music video provider, and Warner Music, a major music company with interests in recorded music, music publishing, and artist services.

%------------------------------------------------------------------------------%

\subsection{Piano Sheet Collection}

For piano MIDI, at first, we try to use optical music recognition for music sheet auto-generation.
However, unlike pure-piano classing music, since most pop music songs are polyphonic, the performance of optical music recognition is imprecise.

Instead, we collected piano scores from MuseScore\footnote{\url{https://musescore.com}}, a popular free and open-source score-writer software, with a website containing more than 1M scores shared by more than 200k musicians.
With such a large community, we can easily find great music for our research.

With some exploration, we found an author named ZakuraMusic\footnote{\url{https://musescore.com/zakuramusic}}, who has more than 11.3k followers and shared 257 music scores (as of the time of this writing).
Most scores are piano rearrangements of famous pop music songs.
The high quality of these scores' BPM makes the alignment much more straightforward.

%------------------------------------------------------------------------------%

\subsection{Music Alignment}

Music alignment is the biggest issue in dataset collection.
Without a perfect alignment, it is not possible to teach our model the relationship between the musicality and the emotion in the video.
In the following, we show some challenges we faced in the alignment:

\paragraph{Most pop music songs do not provide their BPM.}
To deal with this problem, we first find the average BPM of each video, and then generate a MIDI from the piano sheet with the same BPM.
Next, we use OpenShot\footnote{\url{https://www.openshot.org}} for rearrangement.
By overlapping the original video with our MIDI, we can manually check if the correctness of the alignment, and find the arranged measurements in the video.

\paragraph{The BPM in performance is usually not constant.}
Since humans are not perfect metronomes, the speed in their performance is usually not constant.
Fortunately, most pop music songs are adjusted before release.
For those videos with non-constant BPM, we find the BPM of each musical phrase, and either modify the speed of the video or the piano sheet.

\paragraph{The songs in videos differ from the original ones.}
In music videos, it is common to remix the video by adding or removing some fragments.
In this case, we either remove the added fragment from the movie or modify the piano sheet to fit the movie.
Sometimes in music movies, there are tempo changes such as explosions, slow motion clips, or dream fantasy clips.
By using accelerato (acceleration) and rallentando (slow down) in the piano sheet, we may also create these effects in MIDI.
There are also music pauses in some music movies.
These pieces are usually narrating or dialogues, which are not relevant to the music.
Therefore, removing these pieces should not affect the quality of the music generation.

%------------------------------------------------------------------------------%

\subsection{Dataset Summary}

\begin{table}[t!]
\centering
\begin{tabular}{ *{5}{c} } \toprule
%                   &               &               &               &               \\
                    & Songs         & Fragments     & Length (hrs)  & Notes (k)        \\ \midrule
\textit{Training}   & $90$          & $1{,}741$     & $4.93$        & $115.0$       \\
\textit{Validation} & $10$          & $198$         & $0.56$        & $15.5$        \\
\textit{Testing}    & $28$          & $587$         & $1.66$        & $39.9$        \\ \midrule
\textbf{Total}      & $128$         & $2{,}526$     & $7.16$        & $170.5$       \\ \bottomrule
\end{tabular}
\caption{The number of music video songs, 10-second fragments, music lengths, and piano nodes in dataset.}
\label{tab:example}
\end{table}

After the above works, we collected 128 music videos and the corresponding music sheets.
We split this dataset into training (90 songs), validation (10 songs), and testing (28 songs).
Due to the CUDA memory limitation, we divided each song into 10-second fragments.
As show in \tabref{tab:example}, we provide a video to piano MIDI dataset, containing $2{,}526$ samples extracted from $128$ pop music videos.
Each sample contains 10-second fragment of a music video, with $40$ frames scaled to $128 \times 128$ images and rearranged piano MIDI.
The total length of this dataset is $7.16$ hours, with more than $170$k notes.

%% file: content/3-model.tex
%!TEX root = ../vmt.tex

\section{Proposed Model} \label{sec:model}

We describe the proposed video-music transformer (VMT) in this section.
Our approach is inspired by video captioning and text-to-video generation~\cite{chen2015microsoft, venugopalan2015sequence, sun2019videobert}, but deviates from its text generation framework.

\begin{figure*}[t!]
 \includegraphics[width=\textwidth]{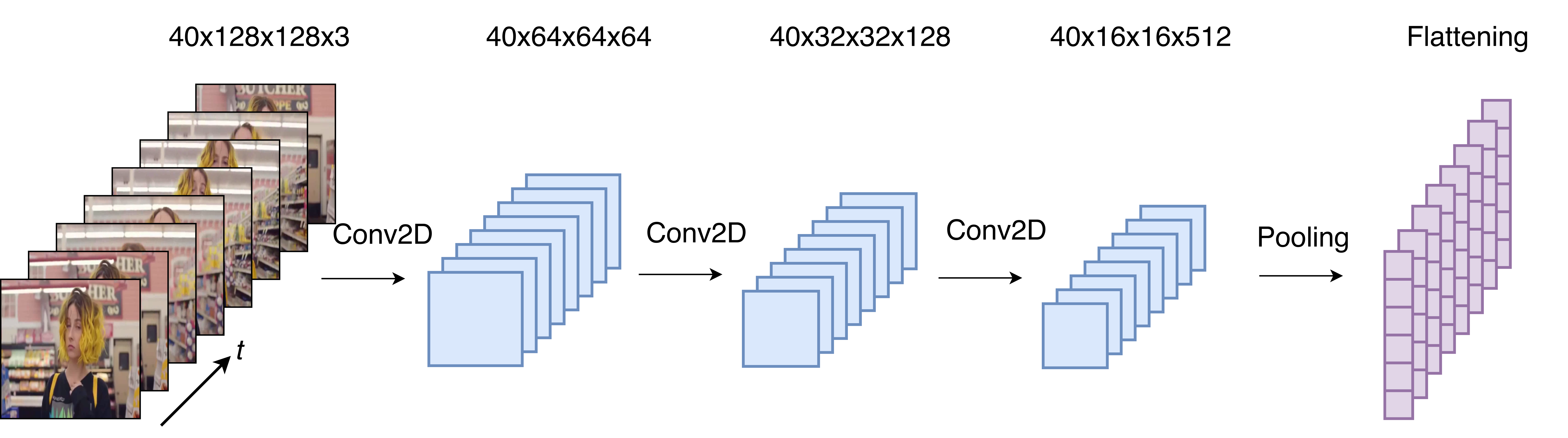}
 \caption{An illustration of video encoding in our models. First, due to the limitation of CUDA memory, we extract equidistantly 40 frames of a video and then reduce the size to $128 \times 128$. Second, we use 3-layer convolutional 2D and an average pooling layer to encode each frame and flatten them.}
 \label{fig:conv}
\end{figure*}

Unlike text sequences, a musical sequence consists of recurring phrases, called a motif, which is of special importance or is characteristic of a composition.
The occurrence of motifs can be sparse. 
Furthermore, musicians often include improvisations to create surprise and variations between performances.
Hence, it is required of our model to maintain long-range ``coherence's''.
Intuitively, an end-to-end attention-based model can be an effective approach to realize this ~--- the attention mechanism creates weighted representation of frame sequences ~--- which then are used to predict piano notes.

In \secref{ssec:model-conv2d}, we show the structure of the convolutional layers in detail. The convolution layers extract and encode the video, including abstract features such as emotion and movement.
\secref{ssec:model-baseline} shows the baseline model to which our VMT will be compared to in the experiments.
Finally, we formulate attention mechanisms such as dot-product (self-attention) and the encoder-decoder attention to predict piano notes in \secref{ssec:model-vmt}.

\paragraph{Notation and Task Definition.} We denote $X=(x_{1}, x_{2}, \dotsc, x_{|X|})$ as a sequence of video frames, where $|X|$ is the length of this sequence.
Each $x_{t}$ consists of three channels of RGB data and encoded with PNG using TensorFlow. The size of $x_{t}$ is $(128, 128, 3)$.
The target of our video-to-music task is a piano event sequence $Y=(y_{1}, y_{2}, \dotsc, y_{|Y|})$.
To generate symbolic music like language modeling, we use the performance encoding proposed by~\cite{oore2018time} to represent the MIDI data.
Hence, Each piano event $y_{t}$ is included in the vocabulary $\mathcal{V}$, $y_{t}\in\mathcal{V}$. Besides,  because of all piano music in our dataset, the vocabulary size $|\mathcal{V}| = 310$.

In this work, our model takes $X$ as input, and outputs sequence of piano event probability $Y$.

%------------------------------------------------------------------------------%

\subsection{Convolutional-2D Encoding} \label{ssec:model-conv2d}
As shown in \figref{fig:conv}, for each frame $x_{t}$, we use a 3-layer 2D convolution with LeakyReLU activation, layer normalization~\cite{ba2016layer} and zero padding given by \eqref{eq:1}, where $x_{t}$ be convoluted with the kernel matrix $k$, with stride $S=2$ and filters $F=64\cdot2^{\frac{1}{2}i(i+1)}, i \in \{ 0, 1, 2\}$. Finally, the output $\hat{c_{t}}$ passes through an average pooling layer to get a flattened embedding vector $a_{t}$ with dimension $H$ which represents one frame.

\begin{equation}
\begin{split} \label{eq:1}
    c_{t}[i, j] &= \sum_{m=-\infty}^{\infty}\sum_{n=-\infty}^{\infty}k[m, n] \cdot x_{t}[i-m, j-n]  \\
    \hat{c_{t}} &= \LayerNorm(\LeakyReLu(c_{t})) \\
    a_{t} &= \AvgPool(\hat{c_{t}})
\end{split}
\end{equation}

%------------------------------------------------------------------------------%

\subsection{Baseline Model} \label{ssec:model-baseline}

\begin{figure}[t!]
    \centering
    \includegraphics[width=\columnwidth]{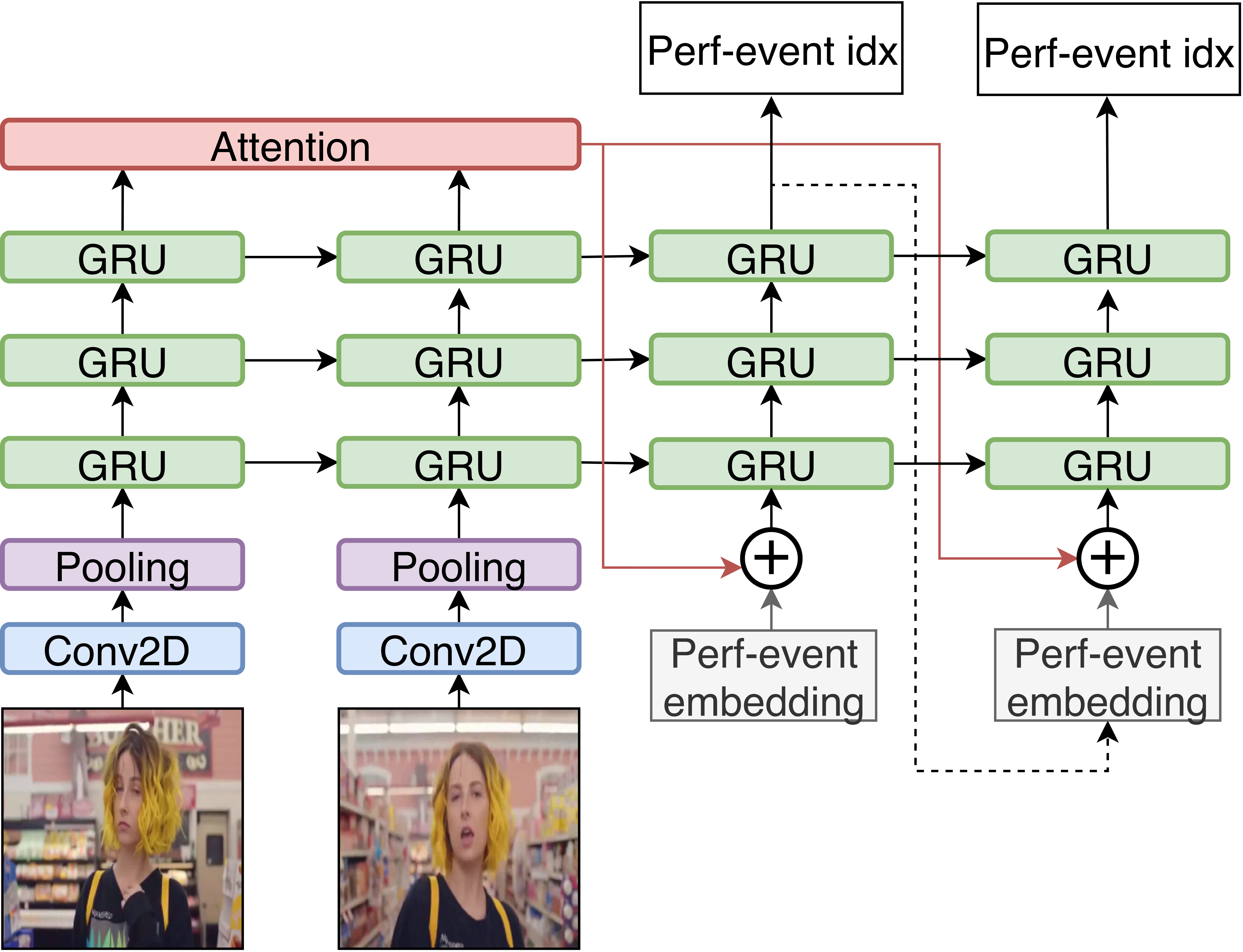}
    \caption{\SeqSeq with encoder-decoder Attention.}
    \label{fig:seq2seq}
\end{figure}

The task of video-to-music conversion is typically viewed as a sequence-to-sequence problem.
Most competitive multi-modal sequence-to-sequence models, including video captioning~\cite{venugopalan2015sequence} and language translation~\cite{bahdanau2014neural, cho2014learning} utilizes an encoder-decoder structure.
Thus, we implement an encoder-decoder model with GRU layers as our baseline model.
As shown in \figref{fig:seq2seq}, we feed the frame vectors $a$ into the encoder. Both the  encoder and decoder architecture are 3-layer gated recurrent units (GRUs). The GRU has a reset gate $r$, update gate $z$ and new gate $n$, which formula are given by \eqref{eq:seq}. The reset gate $r$ and update gate $z$ help the model to determine how much information from the hidden state $h_{(t-1)}$ needs to be passed on.
\begin{equation}\label{eq:seq}
    \begin{split}
        r_{t}&=\sigma(W_{ir}a_{t}+b_{ir}+W_{hr}h_{(t-1)}+b_{hr})    \\
        z_{t}&=\sigma(W_{iz}a_{t}+b_{iz}+W_{hz}h_{(t-1)}+b_{hz})    \\
        n_{t}&=\tanh(W_{in}a_{t}+b_{in}+r_{t}*(W_{hn}h_{(t-1)}+b_{hn})) \\
        h_{t}&=(1-z_{t})*n_{t}+z_{t}*h_{(t-1)}
    \end{split}
\end{equation}
Specifically, we use the same GRU layer for each layer in encoder and decoder. Benefiting from this modification, the decoder is capable of carrying on the information from frame sequence.
Also, we use encoder-decoder attention, which is the same as inter-attention in VMT described in section 3.3, to weight the output hidden state $h^{enc}$ from the encoder in order to get representation as to the input of the decoder. Finally, we use the softmax function on the decoder output $h^{dec}_{t}$ to get the probability of performance event $\hat{y_{t}}$. For the training, we use the negative log-likelihood loss as our objective function.

%------------------------------------------------------------------------------%

\subsection{Video-Music Transformer} \label{ssec:model-vmt}

\begin{figure}[t!]
    \centering
    \includegraphics[width=\columnwidth]{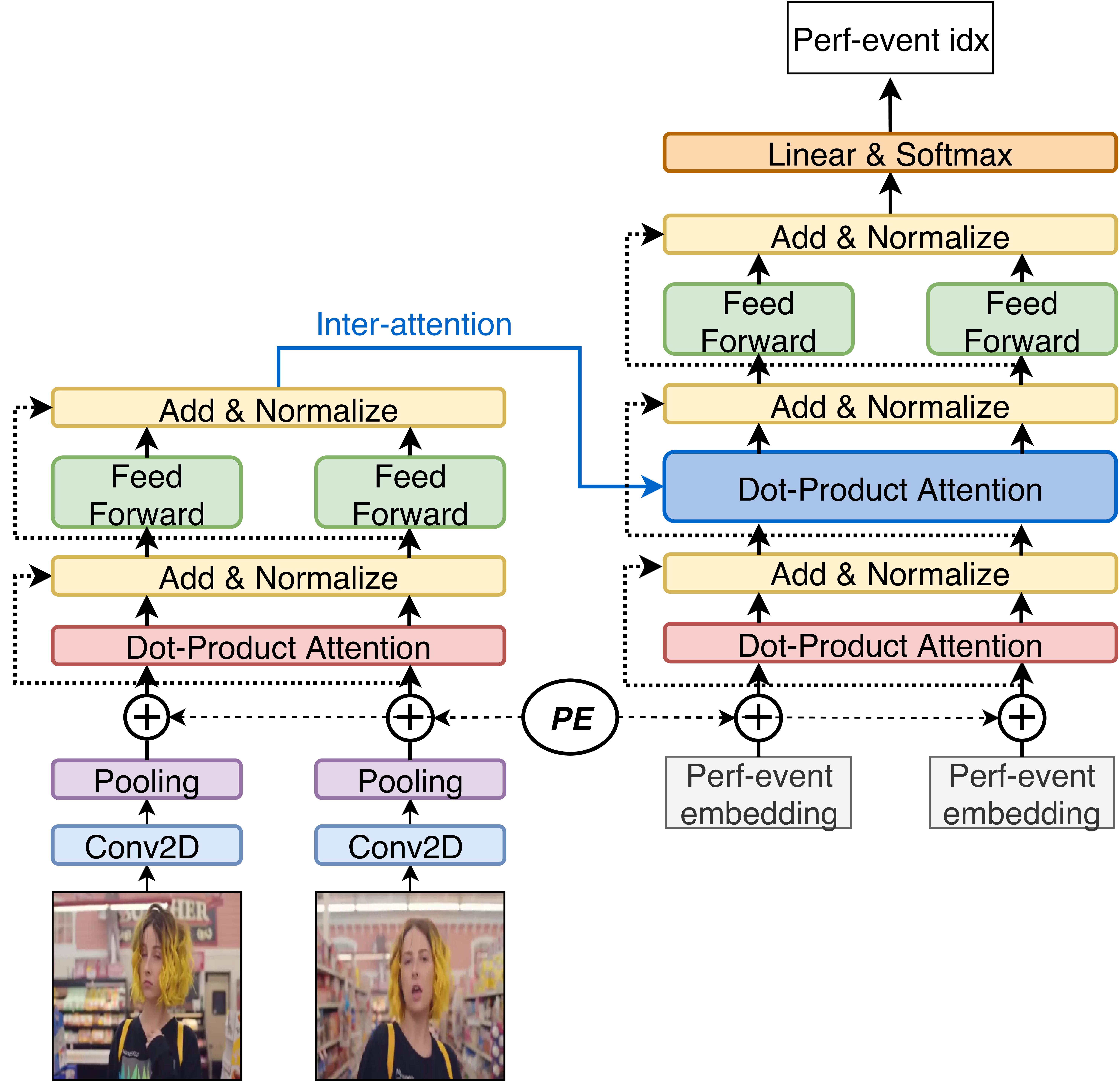}
    \caption{An illustration of the Convolutional Video-Music Transformer.
There are an intra-attention and FFN in each encoder layer, while two types of attention and FFN existing in each decoder layer. Both sequences of hidden frames and event embedding added the positional encoding (PE).}
    \label{fig:vmt}
\end{figure}

As shown in \figref{fig:vmt}, the model architecture of VMT consists of four modules: the 2D convolution encoding, the performance event embedding layer, the encoder and the decoder.
The details of 2D convolution encoding are given in \secref{ssec:model-conv2d}, where frames $X$ are encoded separately to vectors $a=(a_{1}, \dotsc, a_{|X|})$.
The frame vectors $a$ are then fed into dot-product attention. Also, in order to make use of the order of frame sequences, we add positional encodings~\cite{vaswani2017attention} to the frame vectors $a$ and the performance event embeddings $e$ at the bottom of the encoder and decoder, respectively. In this work, we use sine and cosine function of different frequencies:
\begin{equation}\label{eq:pos}
    \begin{split}
        PE_{(pos, 2i)}&=\sin(pos/10000^{2i/H})   \\
        PE_{(pos, 2i+1)}&=\cos(pos/10000^{2i/H})
    \end{split}
\end{equation}
For the performance event embedding $e=(e_{1}, \dotsc, e_{t-1})$ from time step $1$ to $t-1$, we use an embedding matrix $E_{p}\in \mathbb{R}^{H\times |\mathcal{V}|}$, where $H$ is the dimension of the event vector. The values of $E_{p}$ are learned during training.

For each layer in the encoder and decoder, we use dot-product attention and feed-forward network in each layer.
Specifically, since there are different ways to calculate query, key and value, there are two different dot-product attention in our model, called intra-attention and inter-attention.
In every time step, the intra-attention weights different positions of frame sequence in order to compute a representation of video, i.e., $Q=a\times W^{Q},K=a\times W^{K},V=a\times W^{V}$ in the encoder, as well as $Q=e\times W^{Q},K=e\times W^{K},V=e\times W^{V}$ in the decoder.
On the contrary, the inter-attention weights different positions of encoder output vectors, i.e., $Q=z^{enc}_{t}\times W^{Q},K=z^{enc}_{t}\times W^{K},V=z^{dec}_{t-1}\times W^{V}$, where we denote the output of the encoder for each time step by $z^{enc}_{t}$ and $z^{dec}_{t-1}$ indicate the hidden state in the decoder.
Moreover, the hidden state $z^{dec}_{t}$ from inter-attention relates the information from encoder and performance events from $1$ to $t-1$ steps.
Either the outputs $z^{enc}_{t}, z^{dec}_{t-1}$ from intra-attention or the output $z^{dec}_{t}$ from the inter-attention is passed to the feed-forward network.
\begin{equation} \label{eq:dotattn}
    z = \Attention(Q,K,V)
      = \SoftMax \left( \frac{QK^{T}}{\sqrt{d_{k}}} \right) V
\end{equation}

\begin{equation}
    \FFN(z) = \max(0, zW_{1}+b_{1})W_{2}+b_{2}
\end{equation}
Finally, we then pass the hidden state $\FFN(z^{dec}_{t})$ in the last layer to a linear layer and softmax function. The output $\hat{y}$ is the probability of generating a performance event. We compute the negative log-likelihood loss for the target event $y_{t}$ as our objective function.

%% file: content/4-experiment.tex
%!TEX root = ../vmt.tex

\section{Experiment}

\begin{figure*}[t!] 
\centering
\subfloat[][First example by \SeqSeq]{\includegraphics[width=.3\textwidth]{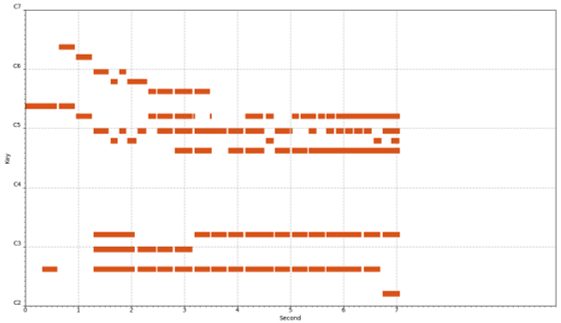} \label{fig:seq1}}
\subfloat[][First example by VMT]{\includegraphics[width=.3\textwidth]{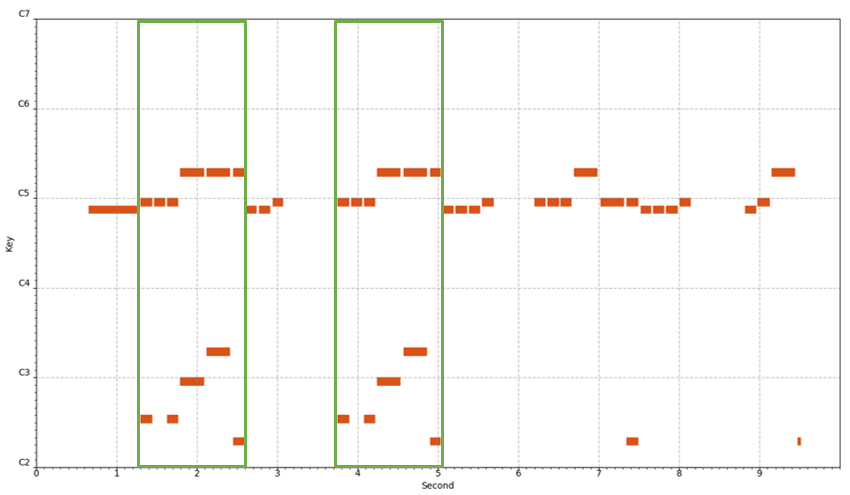} \label{fig:vmt1}}
\subfloat[][First example from original soundtrack]{\includegraphics[width=.3\textwidth]{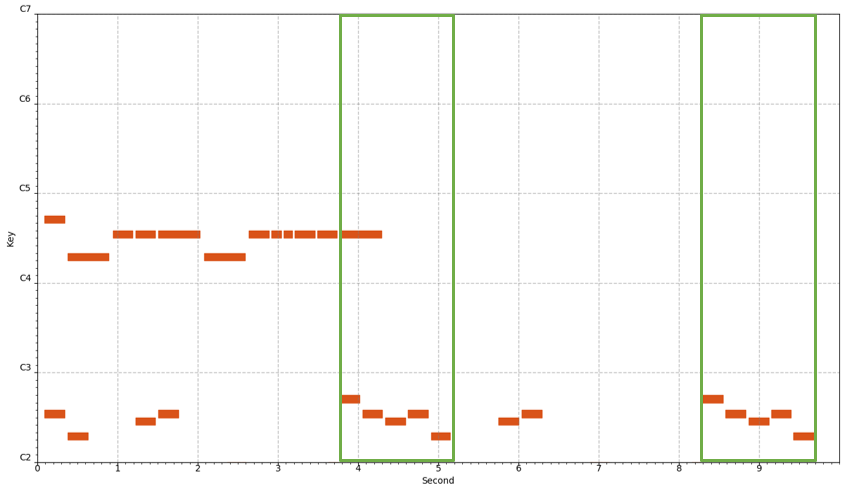} \label{fig:gt1}}

\subfloat[][Second example by \SeqSeq ]{\includegraphics[width=.3\textwidth]{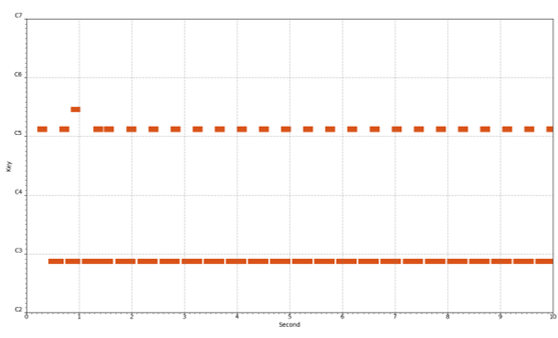} \label{fig:seq2}}
\subfloat[][Second example by VMT ]{\includegraphics[width=.3\textwidth]{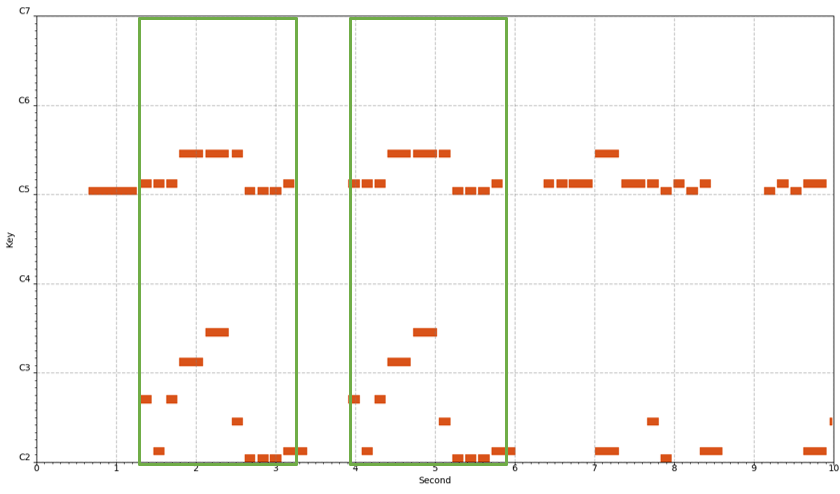} \label{fig:vmt2}}
\subfloat[][Second example from original soundtrack ]{\includegraphics[width=.3\textwidth]{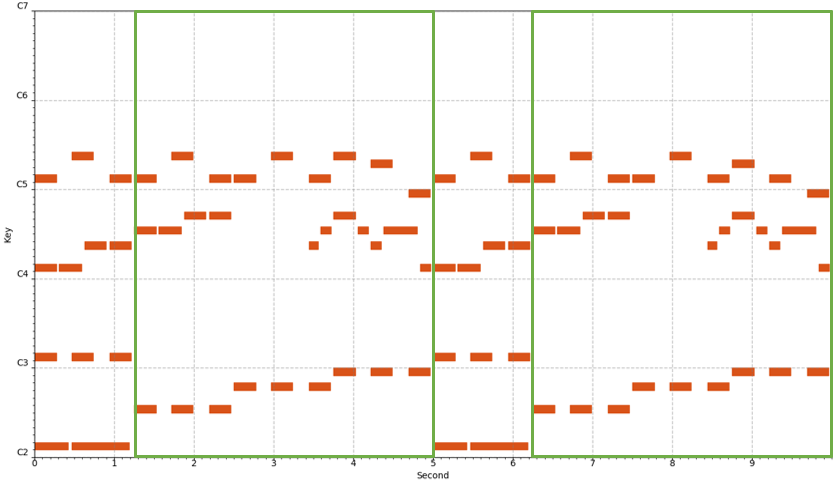} \label{fig:gt2}}
\caption{Two examples individually generated by (a,d) \SeqSeq and (b,e) VMT compared with (e,f) the original soundtrack.
The first example (a) shows \SeqSeq is incapable of generating consistent length with video. The second example (d) shows \SeqSeq tends to generate notes repeatedly and monotonously. By contrast, both examples (b,e) generated by VMT are similar to the original soundtrack. The green vertical boxes indicate a recurring phrase of notes presenting a motif alike. VMT learned to generate melodic and harmonic music.}
\label{fig:three graphs}
\end{figure*}

In this section, we describe the experiment setting and report the results.

%------------------------------------------------------------------------------%

\subsection{Experimental Setting}
We train the VMT model and the \SeqSeq model on a single NVIDIA P40 GPU. Due to the memory limitation, we down-sampled the video by extracting 40 frames from every 10 seconds of video with OpenCV and reduce frame size to $128 \times 128$.
As mentioned previously, our dataset consists of aligned pair of video and piano music, each has the same length of 10 seconds.
We denote the hidden layers and dimensions as $L$ and $H$, respectively. For details, \SeqSeq is trained with three hidden layers.
Both the encoder and decoder in VMT are trained with six hidden layers; additionally, all hidden states have dimension 512, i.e., $L_{seq2seq}=3$, $L_{vmt}^{enc}=6$, $L_{vmt}^{dcc}=6$, $H=512$.
Moreover, we keep a dropout of 0.1 on all layers and attention weights for \SeqSeq and VMT model. The learning rate is linearly warmed up over the first 8000 steps to achieve a peak value of $10^{-3}$ and then decayed with the inverse square root of the steps.
For the optimizer, we use Adam with hyperparameters $\beta_{1}=0.9$, $\beta_{2}=0.997$ and $\epsilon=10^{-9}$, which converges to better sets of our model parameters. Our implementation uses a batch size of 4 video frame sequences, whose size is $40 \times 128 \times 128 \times 3$. We train the VMT model for a total of 50,000 steps over 18 hours.

In particular, the maximum length of the target sequences of VMT is set to 1024. Since we use the performance encoding on our target piano notes, the vocabulary size of events is 310, including 88 \verb|NOTE_ON|, 88 \verb|NOTE_OFF|, 32 \verb|TIME_SHIFT|, 100 \verb|VELOCITY|, and two special tokens represent the start and end of event sequences.

%------------------------------------------------------------------------------%

\subsection{User Evaluation}
In this paper, we are working on a new task of video-to-music translation, and there is yet a credible benchmark for this task.
Therefore, we compare the proposed model VMT with \SeqSeq and original music from the testing dataset.
We conduct a user evaluation involving 23 participants, with each participant evaluating 30 randomly sampled examples from the testing set.
In each trial, the participant is asked to compare three pieces of music accompanying the same video. The pieces included one generated by VMT, one by \SeqSeq and the original soundtrack. 
Participants compare the three pieces and score each piece by the smoothness of the music and it's relevance to the video. Each score ranges from 0 to 5.
Since our dataset is collected from pop music videos, we design a button ``I know this song or music video. Skip!'' to avoid biases from affecting the accuracy of evaluation.
For example, the songs of ``Adele~--- Hello'', ``Ed Sheeran~--- Shape of you'' and ``Lady Gaga~--- Poker face'' achieved high skip rates compared with the other samples in the testing set for the experiment.
We report in \tabref{tab:user} the scores that the two models and the original music receives.

\begin{table}[t!]
\centering
\begin{tabular}{ *{4}{c} } \toprule
%                      &               &               &               \\
                       & Original      & VMT     & \SeqSeq       \\ \midrule
\textit{Smoothness}    & $3.60$        & $3.11$        & $2.74$        \\
\textit{Relevance}     & $3.05$        & $2.36$        & $2.13$        \\ \bottomrule
\end{tabular}
\caption{Result of user evaluation, the smoothness is defined as the qualities in pieces of music that sound melodiously and harmoniously. The relevance is defined as the music is a proper soundtrack for the video. The column ``original'' refers to the original soundtracks.}
\label{tab:user}
\vspace*{-1em}
\end{table}

% Result of user evaluation, the smoothness is defined as the qualities in pieces of music that sound melodiously and harmoniously. The relevance is defined as the music is a proper soundtrack for the video.
As shown in \tabref{tab:user}, although the original soundtracks remain far ahead in terms of both smoothness and relevance scores, the proposed VMT model has substantially outperformed the baseline seq2seq model.
Each score in \tabref{tab:user} is the average value of the total of 690 samples graded from 23 participants.
Interestingly, both generated music and original soundtrack whose scores for the relevance of video are lower than the scores of music smoothness.
Since our dataset consisted of pop music videos and pop music. It is possible that limiting the model to generate only piano MIDI impedes the user experience of watching the video and consequently hurts the relevance performance.

%------------------------------------------------------------------------------%

\subsection{Case Study}
We choose two samples from the testing set and visualized the MIDI files generated from both VMT and \SeqSeq, compared with the original soundtracks. \figref{fig:three graphs} shows the notes played along with seconds, i.e., the x-axis and y-axis are pitch and seconds, range from C2--C7 and 0--10, respectively. 

Taking a closer look at \figref{fig:seq1}, \ref{fig:seq2}. 
While samples generated with \SeqSeq achieve higher scores on smoothness and relevance, there are two critical problems for them.

Firstly, as shown in \figref{fig:seq1}, even though the length of all videos is ten seconds, \SeqSeq is incapable of generating music whose length is exactly ten seconds.
That is, \SeqSeq fails to correctly generate end tokens, resulting in an inconsistency between the music and the video. We removed the music pieces over ten seconds for the user evaluation. 
Moreover, while decoding the output event sequence from the \SeqSeq model using the performance encoding module, there are some warning messages such as a pitch with \verb|NOTE_ON| events but not found \verb|NOTE_OFF| would be removed.
Thus, we suppose that the selected metrics have overrated samples generated with \SeqSeq. And verify the \SeqSeq is a lack of maintaining long-term sequence generation. 
In contrast, the examples generated with VMT (\figref{fig:vmt1}, \ref{fig:vmt2}) have consistent lengths with videos and without warning message while decoding using the performance encoding module.

Secondly, we observe the rhythmicity and harmonicity of note sequences.
\SeqSeq tends to generate notes repeatedly (\figref{fig:seq2}), which causes a lack of melodic motion.
This example gets a higher score on relevance, since the user mistook the generated notes for the beat of drums. This also demonstrated that the \textbf{relevance} metric of the video is ambiguous and needs a benchmark to train users before doing the evaluation.

% A new subsection maybe?
As shown in \figref{fig:three graphs}, the green vertical boxes indicate a recurring phrase of notes. Both \figref{fig:vmt1} and \figref{fig:vmt2} generated with VMT has a motif, which is the most often thought of in melodic terms. We observe the VMT not only generates recurring phrases but also maintains the variability of pitches that have harmonic, melodic and rhythmic aspects.

In summary, we compare examples from models with the original soundtrack (\figref{fig:gt1}, \ref{fig:gt2}). The visualization (\figref{fig:vmt1}, \ref{fig:vmt2}) shows the VMT is capable of generating notes with the recurring phrase, melodic and harmonic. Hence, the training procedure passes knowledge of music structure onto the VMT model and composes music like a human.

%% file: content/5-related.tex
%!TEX root = ../vmt.tex

\section{Related work}

The closest task to video-to-music is the music recommendation for video. There are two approaches for music recommendation: emotion-based and correlation-based. The first emotion-based model~\cite{kuo2005emotion} focus on user emotions, which conduct a mixed media graph to detect music emotion rather than directly using the label of music emotion.

On the other hand, the EMV-matchmaker~\cite{lin2015emv} proposed to extract video and music features separately and then utilize the temporal phase sequence to connect music and video.
Before that, most of the researches on video music recommendation based on the similarity of user preference.

Music video generation~\cite{lin2016automatic} combined the content and emotion features.
Given a video, their model predicts the acoustic features to match the music.

To sum up, those works are video retrieving music task, that may suffer from the copyright and the insufficient diversity of database. Moreover, the music generated by the model are more varied and fitting the video.

With the advancement in deep learning, music generation models has improved dramatically. 
Some works developed depending on symbolic music, that is, the target data of training are MIDI files.
The first neural-network-based model on music generation was proposed by~\cite{todd1989connectionist}. 
Along with the development of model structure, many music generation models are based on RNN  architectures~\cite{hadjeres2017deepbach, jaques2016generating}, due to the sequential nature of the input.

C-RNN-GAN~\cite{mogren2016c} is the first model based on adversarial training.
The MIDI-VAE~\cite{brunner2018midi} based on variational autoencoder is capable of style transfer on symbolic music by changing pitches and instruments.
A branch of multi-track music generation studies appeared along with the release of the Lakh MIDI Dataset~\cite{raffel2016learning}.
The LakhNES~\cite{donahue2019lakhnes} was a transformer architecture model with a pre-training technique to generate chiptune music.
MusuGAN~\cite{dong2018musegan} generated polyphonic music of multi-track instruments, which using convolutions in both generators and the discriminators, which conditioned by intra-track and inter-track features.

On the other hand, following the generative model of raw audio waveforms~\cite{oord2016wavenet}, an autoencoder model~\cite{engel2017neural} is trained to learn music features and generate music mixed with bass, flute, and organ spectrum.
The music synthesis model Mel2Mel~\cite{kim2019neural} learned the instrument embedding to predict Mel spectrogram from a given note sequence.
The release of MAESTRO~\cite{hawthorne2018enabling} enabled the process of transcribing, composing, and synthesizing audio waveform, also known as Wave2Midi2Wave.
The model~\cite{engel2019gansynth} trained on the NSynth dataset~\cite{engel2017neural} was capable of independently controlling pitch and timbre then generate audio music.

%% file: content/6-conclusion.tex
%!TEX root = ../vmt.tex

\section{conclusion and future work}

We propose a novel attention-based multi-modal model, video-music transformer, called VMT, which generates piano music for a given video. We release a new dataset composed of over 7 hours of piano scores with excellent alignment with video. The experiment shows VMT outperforms the \SeqSeq model on music smoothness and relevance of video.
In future work, we plan to explore proper benchmark of relevance with video, including emotion, rhythm, and motion connection. Furthermore, we will develop the model architecture to encourage the model to learn these features.